\documentclass{article}

\usepackage[letterpaper,top=2cm,bottom=2cm,left=3cm,right=3cm,marginparwidth=1.75cm]{geometry}

\usepackage{amsmath}
\usepackage{graphicx}
\usepackage[colorlinks=true, allcolors=blue]{hyperref}
\usepackage{booktabs}

\author{
  Raul Gomez Bruballa\\
  Shutterstock
  \and
  Lauren Burnham-King\\
  Shutterstock
    \and
   Alessandra Sala\\
   Shutterstock
   \and
   \texttt{\small\{rgomezbruballa\}\{lburnhamking\}\{asala\}@shutterstock.com}
}

\date{}
\title{Can you recommend content to creatives instead of final consumers? A RecSys based on user's preferred visual styles \\
\begin{large} 
  Technical Report
\end{large}}

\begin{document}
\maketitle

\begin{abstract}
Providing meaningful recommendations in a content marketplace is challenging due to the fact that users are not the final content consumers. Instead, most users are creatives whose interests, linked to the projects they work on, change rapidly and abruptly. 
To address the challenging task of recommending images to content creators, we design a RecSys that learns visual styles preferences transversal to the semantics of the projects users work on. 
We analyze the challenges of the task compared to content-based recommendations driven by semantics, propose an evaluation setup, and explain its applications in a global image marketplace.   

This technical report is an extension of the paper \textit{Learning Users' Preferred Visual Styles in an Image Marketplace}, presented at \textit{ACM RecSys '22}.
\end{abstract}

\section{Introduction}

Content marketplaces are huge and diverse online catalogs that users navigate leveraging search engines, which rank content based mainly on popularity, to find appropriate assets for their creative projects.
Nowadays, those marketplaces constitute a key element in almost any content creation process, from an advertisement poster, to a news video or an educational book. Consequently, the content they serve has a huge influence in any media creation.

Image marketplaces are global, but people are diverse in terms of visual preferences which are influenced by the people's culture, age, and entire life experiences. 
Given that image marketplaces are the first stage in many visual content generation processes, it is crucial to maintain that diversity in the images they serve, since that can help to prevent that majority style preferences (and ergo cultural preferences) prevail. 
In this work, we propose leveraging a RecSys to personalize the images served to each user in an image marketplace, which helps creatives finding relevant images for their projects and boosts the diversity of the content served.

RecSys are a great content discovery tool to assure diversity, since they can personalize the content served to each user.
However providing meaningful recommendations in an image marketplace is a challenging task because platform users are not the final content consumers but creatives whose interests shift rapidly with the projects they work on. 
An option to address those rapid interests shifts would be to provide recommendations based on the latest user activity; i.e. recommendations based on users' latest search queries. 
However, those recommendations wouldn't be personalized for each user itself and therefore wouldn't ensure diversity in the content served. Instead, we aim to explore historical creatives activity in the platform to learn long term preferences which are stable across the projects they work on.

We hypothesize that creatives have visual style preferences that are transversal to the semantics of their projects. Those preferences might have diverse natures: i.e. a user might have preferences for images by Indian photographers and close-up images from Italy, while another one for images of isolated objects with white backgrounds and wide-angle images of landscapes. However, with an adequate feature selection, a RecSys model can learn which style features (or styles defined by features interactions) are relevant for each user.
Marketplaces are rich in terms of images metadata, but designing features that encode creatives' visual style preferences is challenging. We achieve that in collaboration with content experts with a strong knowledge of creatives' behaviours and interests.
Once image features that encode a user's long term creative preferences have been designed, a RecSys is trained with historical platform clicking data to learn users' preferred visual styles. Finally, the content served in the image marketplace is personalized for each user, ensuring it matches creatives' visual styles preferences and promoting diversity.

\begin{figure}[t]
  \centering
  \includegraphics[width=\linewidth]{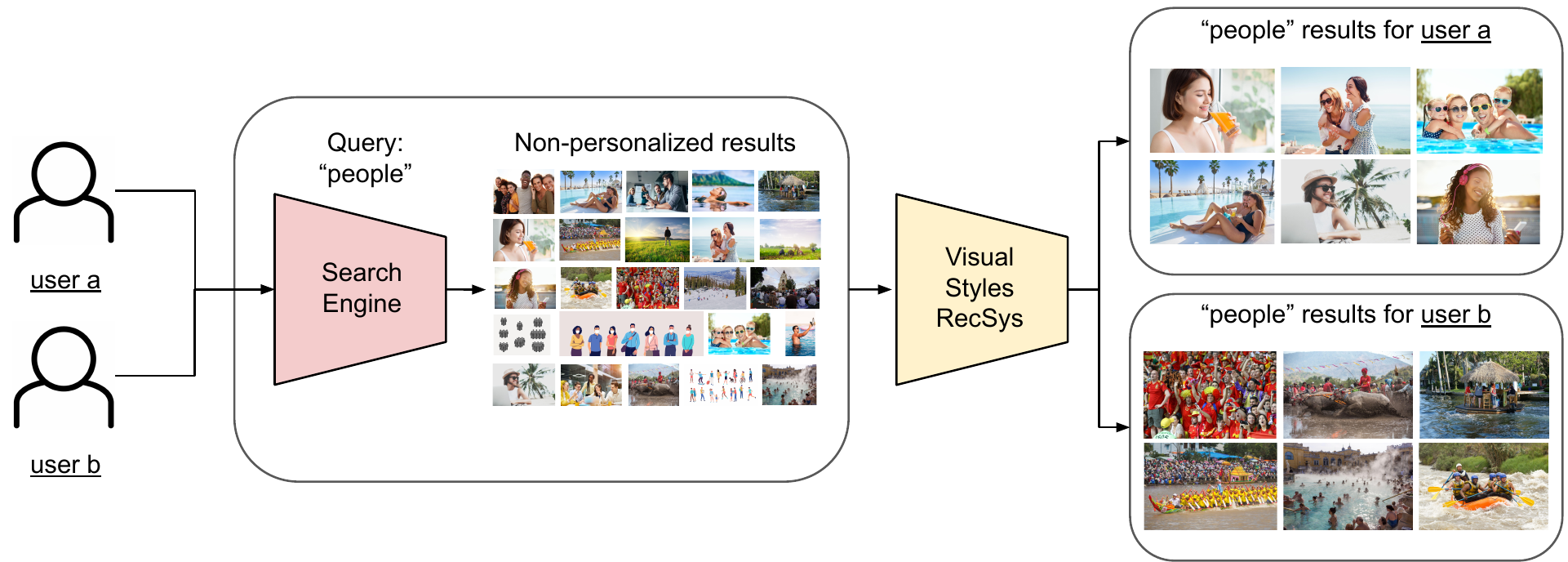}
  \caption{Recommendations pipeline. A user queries the search engine, which returns non-personalized results matching the query semantics (in the example, "people"). Then, Visual Styles RecSys re-ranks those results, and the ones inline with the user preferred visual styles are shown first. Note how in the example, while the search engine provides very diverse results, \underline{user a} gets more stocky portraits with blurred backgrounds, while \underline{user b} gets served first top-view cadid images of crowds. None of the users seems interested in, as an example, illustrations.}
  \label{fig:pipeline}
\end{figure}

\hfill \break
\textbf{Our summarized contributions are:}
\begin{itemize}
\item We investigate the problem of recommending content to creatives who are not the final content consumers.
\item We design an approach that learns visual style preferences which are transversal to creatives' projects semantics.
\item We collaborate with content experts with a strong knowledge of creatives' behaviour for feature selection and qualitative recommendations evaluation and ensure that our approach boosts content diversity.
\item We propose an evaluation framework and perform an ablation study which is proven key to understand the model behaviour and the selected features influence in the recommendations.
\end{itemize}

The paper is structured as follows: First we go though related work and literature of interest (\ref{sec:related_work}). Next we formulate the problem  and the proposed solution for the image marketplace (\ref{sec:problem}). Then we describe the data collection process and the user and image features we experiment with (\ref{sec:data}). Following we describe the RecSys model architecture and optimization setup (\ref{sec:model}), and finally we show experimentation results (\ref{sec:experiments}) and draw conclusions (\ref{sec:conclusions}).

\section{Related Work}
\label{sec:related_work}

The peculiarity of the addressed task is that we aim to learn long term creatives preferences (visual styles) leaving out short term interests (that change rapidly with creatives' projects). 
That differs from the objective in most scenarios, such as recommendations in a video sharing platform \cite{youtube}, a video streaming platform \cite{netflix_search}, or an online goods marketplace \cite{amazon}, where the user is the content consumer and short term and long term preferences are entangled and exploited jointly. There are some works addressing the fusion of long term and short term preferences to boost the latest while benefiting from both, as this method \cite{spotify} doing so for music recommendation. However, there is no literature we are aware of focusing on learning long term preferences.

An straight forward solution to learn long term interests would be to use a big training window so that short term interests are diluted in it, but that would lead us to huge training datasets and add considerable complexity since the optimal training window highly differs among users. Instead, we decided to collaborate with content experts to handcraft image features encoding long term preferences, a process which was key and is necessary in any RecSys involving editorial guidelines \cite{bbc}.
Fomenting diversity and fighting bias is one of the main editorial guidelines in our content marketplace, and it's well known \cite{bias} that RecSys face many bias problems which are challenging to handle. However, the proposed solution has been designed to boost diversity in the images served while being robust to RecSys biases: First, because it personalizes search results at a user-level and, second, because recommendations are based on image style features which, due to their low discriminating power, are less prone to introducing undesirable biases.

This paper focuses on describing a method to learn creatives visual style preferences transversal to their short term interests. It also exploits and benchmarks state of the art modelling solutions, such as Deep Cross Networks \cite{dcn1, dcn2}, and leverages recent evaluation metrics, such as \texttt{Effective Catalog Size} \cite{ecs} to measure recommendations diversity, and LPIPS \cite{lpips} (a metric used to measure the diversity of images generated by Generative Adversarial Networks) to measure recommendations \texttt{Visual Diversity}, which helps monitoring the users' preferences learned by the proposed \texttt{Visual Styles RecSys}.

\section{Problem Formulation}
\label{sec:problem}

\subsection{Recommendations for Creatives}
An image marketplace is a huge catalog of very diverse content in terms of semantics, where one can find from images of a train to images of a surgery, and also in terms of styles, where a train image can be a grayscale illustration or a landscape wide-angle image of a train crossing the Alps.
Platform users are creatives whose content interests change rapidly since are dependent on the different projects they work on, and to navigate the catalog they use a search engine which provides content relevant to a given search query, ranked mainly by popularity.

As the catalog is diverse in terms of visual styles, creatives also are in terms of visual style preferences, and we hypothesise those are stable across projects. As an example, if a creative has licensed mostly illustrations in the marketplace, when he searches for "train" he will be probably interested in train illustrations, while if another user has licensed many wide-angle images of landscapes from Switzerland, when he searches for "train" he will probably be more interested in the train image crossing the Alps.
In the proposed recommendations pipeline, we maintain the ability of creatives to search for content relevant for their projects, but we personalize search engine results to match each user's preferred visual styles, which we learn from their historical activity in the marketplace, aiming to serve content more relevant to them.

A key aspect of the \texttt{Visual Styles RecSys} is therefore to define image features encoding styles which are stable across creatives' projects. To do that we collaborated with the image marketplace content experts analyzing users' behaviour in the platform to get insights of which long term style preferences they have and which image features encode them. 
Features of different natures were found relevant and later tested experimentally. Among them general features as the image predominant colors or author country, features handcrafted from selected keywords as the image angle, or semantic verticals such as the image category. Selected features are described in detail in Section \ref{sec:image_features}.
Note that, for an style feature to be useful for recommendations, it doesn't need to be stable across projects for all users: As explained in Section \ref{sec:model}, the model can learn which features or features interactions are useful to recommend relevant content to each user. 
However, it is critical to avoid features encoding project semantics, because they are too discriminative compared to style features and would prohibit learning users' preferred styles. As an example, if we include image keywords as a feature, the model would rapidly fit the keywords relevant for the projects a user worked on during the training window, instead of learning styles relevant for further projects.

\subsection{Recommendations Pipeline}

The pipeline to personalize content discovery at the image marketplace exploiting \texttt{Visual Styles RecSys} is depicted in Figure \ref{fig:pipeline}. It is a two-stages pipeline: First a search engine generates recommendation candidates relevant to a user search query; Then, the RecSys re-ranks those candidates, so that the ones inline with the user visual styles preferences are served first.

The interest of maintaining the search engine in the recommendation pipeline is that creatives can still benefit from custom recommendations while they rapidly shift between projects. Additionally, they can utilize at the same time any search engine functionality, such as search filters or trending images boosting. 
It is indeed crucial that the search engine provides candidates with enough styles diversity so that they fit the preferred styles of any user, and the number of recommendation candidates used is among hundreds. This paper focuses on the RecSys modelling and leaves the integration with the search engine and applications for further discussion.

\section{Data}
\label{sec:data}

\begin{figure}[h]
    \centering
    {{\includegraphics[width=5cm]{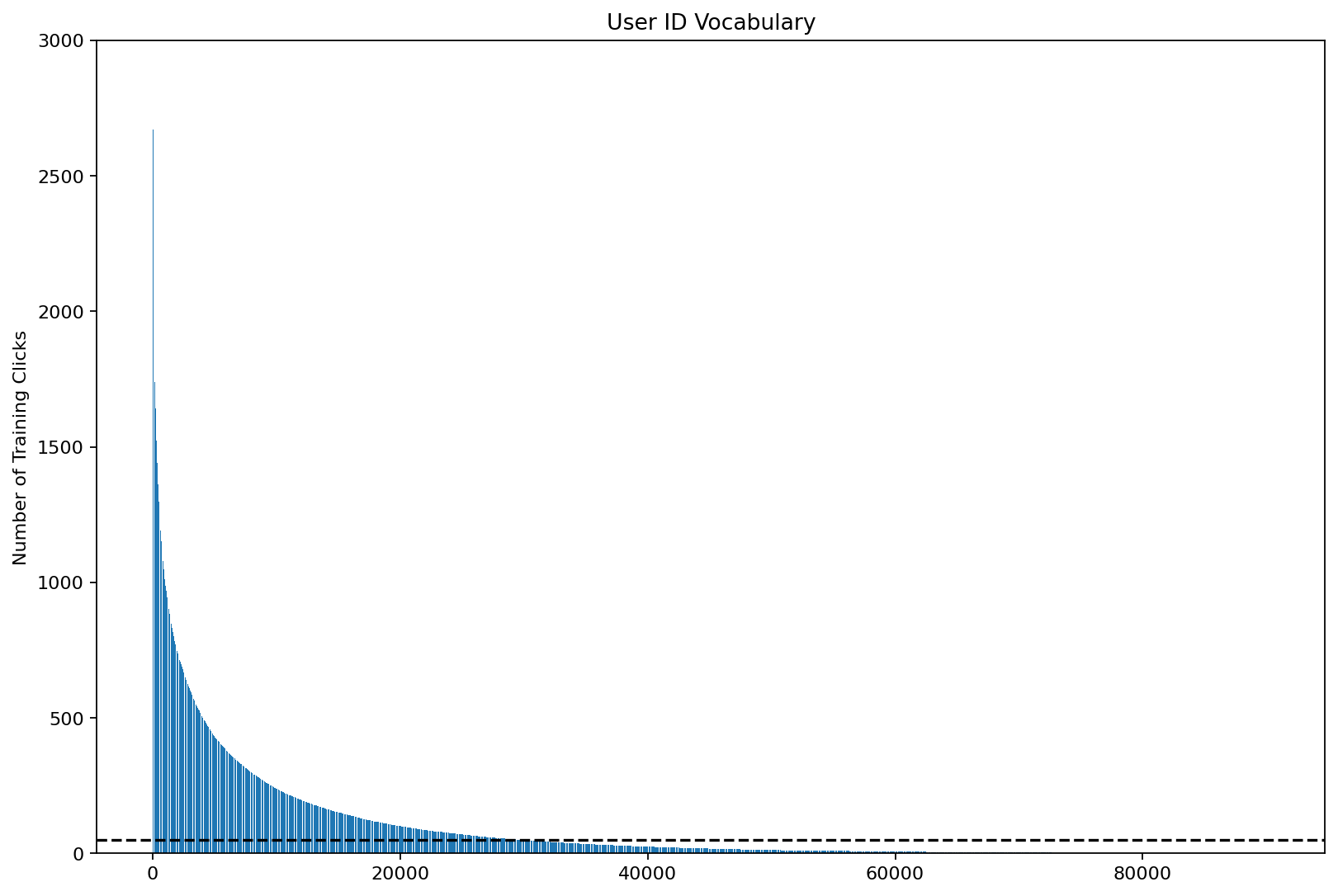}}}%
    \qquad
    {{\includegraphics[width=5cm]{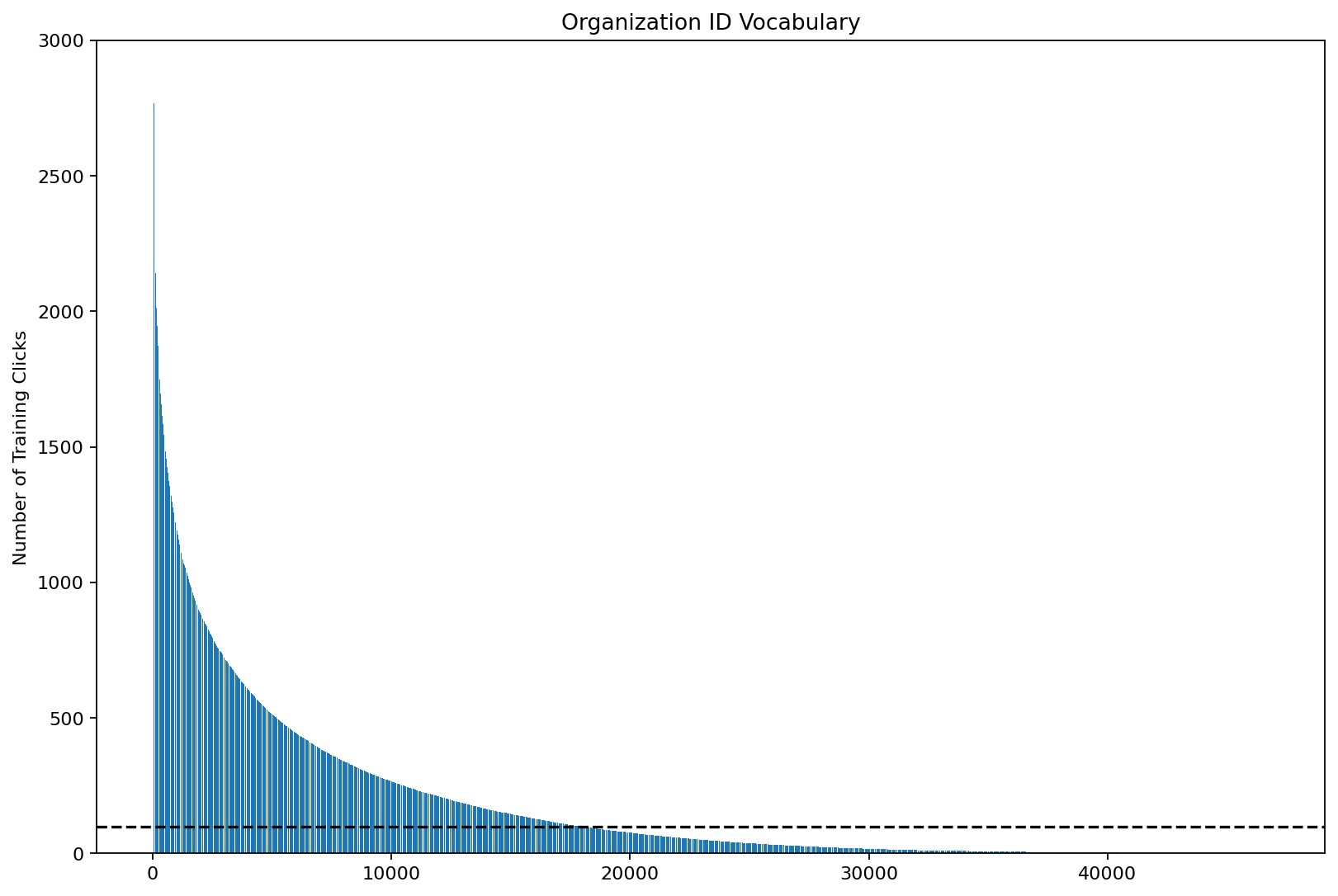}}}%
    \caption{Users' clicks distribution (left) and users' organizations clicks distribution (right). The horizontal lines (at 50 and 100 clicks respectively) indicate which users and organizations are included in the vocabulary and therefore have a unique encoding.}%
    \label{fig:dataset}%
\end{figure}

To design the architecture of the \texttt{Visual Styles RecSys} model we collect a dataset consisting of $10M$ user clicks in images of $90K$ users during a $6$ months time window. We split those in a training set and a testing set, the latter gathering the latest $1M$ clicks. The number of different images is the dataset is $4M$. Figure \ref{fig:dataset} shows that the number of clicks per user and user organization follows a long tail distribution, representing users' behaviour at the marketplace.
For each user and image in the dataset we collect a set of features, which are explained next.

\subsection{User Features}

\begin{itemize}
\item {\texttt{User ID}}: Unique user identifier. For users with more than $50$ clicks ($30K$ users; see Figure \ref{fig:dataset}) is encoded uniquely as a one-hot vector over an IDs vocabulary, and the rest of the user IDs are mapped to one of $10K$ out of vocabulary tokens. The reason to do that is, first, to reduce model size and, second, to even how fast the model fits the style preferences of each user. 
\item {\texttt{Organization ID}}: Unique identifier of the organization the user is affiliated to. All users have an organization, but an organization can consist of a unique user. It's encoded similarly as User ID: Organizations with more than $100$ clicks ($18K$ out of $47K$) are encoded uniquely, while the rest are mapped to one of 10k out of vocabulary tokens.
\item {\texttt{User Country}}: Country of the user encoded as a one-hot vector. Countries with less than $200$ clicks are clipped, resulting in a $142$ countries vocabulary.
\item {\texttt{User Language}}: Language of the user encoded as a one-hot vector. Languages with less than $200$ clicks are clipped, resulting in a $24$ languages vocabulary.
\item {\texttt{Subscription Product}}: User subscription plan at the content marketplace, encoded as a one-hot vector.
\end{itemize}

\subsection{Image Features}
\label{sec:image_features}

\begin{itemize}
\item {\texttt{Categories}}: Images are annotated with $1$ or $2$ of $30$ categories, which are generic verticals such as “Nature”, “Signs/Symbols” or “Illustrations”. They are encoded as a multi-hot vector.
\item {\texttt{Contributor Country}}: Country of the image author encoded as a one-hot vector. Countries with less than $200$ appearances are clipped, resulting in a $146$ countries vocabulary.
\item {\texttt{Colors}}: Image dominant colors computed based on hue histogram color peaks, which are then mapped to one of $138$ HSV named colors. They are encoded as dense vectors, where each value corresponds to the normalized peak coverage, if any, for a given named color.
\item {\texttt{Image Angle Features}}: Categorical features encoding the image angle, based on keywords handcrafted by our content experts team. Examples of angles encoded by these features are “aerial view”, “wide angle view”, or “close up view”.
\item {\texttt{Photo Style Features}}: Categorical features encoding photo styles based on keywords handcrafted by our content experts team. Examples of styles encoded by these features are “candid style”, or “studio isolated style”.
\item {\texttt{Deep Features}}: Visual features extracted by an Inception v1 CNN \cite{googlenet} trained with the marketplace popular search queries results. Dense vectors of $256$ dimensions. We also experiment using CLIP \cite{clip} representations instead.
\end{itemize}


The \texttt{Categories} feature encodes some style preferences, such as “Illustrations”, but also some semantic verticals preferences, such as “Nature” or "Sports/Recreation". We have observed that they are stable across projects for some creatives. For instance, when searching for "train" images, a user with a high preference for “Nature” images might be more interested in images of trains across landscapes, but a user with strong preferences towards "Sports/Recreation" might be interested instead in images of people training. For users without strong categories preferences, recommendations will be more diverse in terms of categories and driven by other style features.

Deep features tend to contain very rich and discriminative visual information. A risk of using them, similarly as we explained with keywords, is that the model could leverage them to end up learning the semantic preferences of users during the training window, instead of learning long term style preferences useful for further projects. 
We propose a metric to measure how much the model relies on deep features to provide recommendations, which shows that the used deep features are compatible with learning long term style preferences. The main reason is that the used features do not encode fine-grained semantics. As a contrast, we experimented using CLIP representations instead, and found that in that case recommendations were solely based on semantics, and therefore irrelevant for future creatives projects. This is measured and discussed in Section \ref{sec:experiments}.

\section{Model}
\label{sec:model}

\subsection{Architecture}

\begin{figure}[h]
  \centering
  \includegraphics[width=\linewidth]{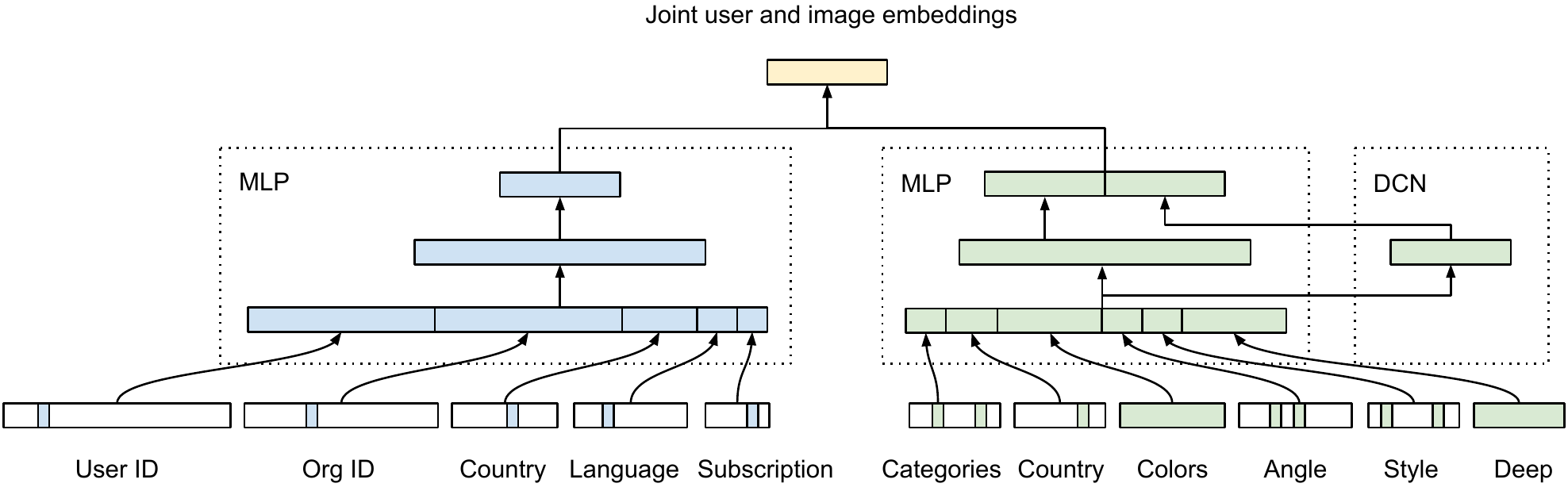}
  \caption{Architecture of the \texttt{Visual Styles RecSys}. In blue, layers of the user encoder and in green, layers the image encoder. MLP (Multi Layer Perception) layers are linear layers with ReLU activations, and DCN (Deep Cross Network) layers are Cross layers (in this case a single Cross layer is used).}
  \label{fig:model}
\end{figure}

\texttt{Visual Styles RecSys} has a two-tower architecture and learns users and images representations in a joint embedding space, as shown in Figure \ref{fig:model}. 
The user encoder is essentially a Multi Layer Perception. First, representations of each one of the user features of experimentally found dimensionalities are learned, then those representations are concatenated and go through a couple of fully connected layers with ReLU activations and finally a linear layer learns the projection to the embedding space, with a dimensionality of $128$. 

In the image tower, representations of each one of the image features are learned and concatenated to be processed in parallel by a MLP and a Cross Network \cite{dcn2} with a single Cross Layer. Finally features coming out from the two branches are concatenated and projected into the embedding space.

Cross Layers \cite{dcn2} explicitly model feature interactions, which seems critical for our RecSys aiming to learn users' preferred styles based on interactions among image style features. We experimented using multiple Cross Layers and placing them before the MLP, but results didn't improve further. We also experimented using them in similar setups in the user tower, and resulted in a similar performance with a lighter model in terms of parameters, but we dropped that solution since despite having less parameters resulted in slower inference.

The two tower architecture facilitates a fast inference setup, since image embeddings for the whole catalog can be computed offline. When a user performs a query, his embedding is computed, recommendation candidates are retrieved, and distances between the user embedding and the candidates embeddings are computed to get the RecSys ranking.
\texttt{Visual Styles RecSys} is built on Keras leveraging the TensorFlow Recommenders library \cite{tfrm}.

\subsection{Training}

The collected dataset gathers data about which images a user has clicked on in the past, and the model is optimized to embed users close to the images they have clicked on (positive images) and far from others (negative images) in the joint embedding space.
Specifically, for a batch of $N$ (user, image) pairs this is implemented maximizing the dot product between a user embedding and the embedding of the image he has clicked on ($N$ pairs), and minimizing it for the rest of the images in the batch ($N^2 - N$ pairs), which are used as negatives. We optimize a symmetric cross entropy loss over these similarity scores. 
This optimization strategy is very efficient and has been proven superior in many tasks, from unsupervised representation learning for different modalities \cite{cpc} to learning joint image and text embedding leveraging text supervision \cite{clip}.

The two towers are trained at once with an Adagrad optimizer, and we found a long warm up to be crucial for a proper optimization, which we attribute to the diverse nature of the input features. The optimal learning schedule found was starting with a learning rate of $1e-5$ and increase it an order of magnitude at each epoch until $1e-2$.
A random $10\%$ of the training set is employed as validation, and models are trained until validation loss convergence.

Selecting an appropriate batch size is critical. A small batch size results in very noisy gradients which dramatically slow down optimization. However, contrary to other tasks with similar optimization setups, using large batch sizes obstructs optimization. That's because in our case the used image style features are not highly discriminative, and by using large batch sizes we increase the number of negatives at each iteration, which makes the task of learning patterns to distinguish between positives and negatives images too challenging. We found a batch size of $256$ to be optimal.

\texttt{Visual Styles RecSys} is trained with data covering a $6$ months time window, and it's intended to learn long term user preferences.
However, those long term preferences might slowly change, so at some point one might want to re-train the model with fresh data to update them. 
Note that, to learn fresh user preferences, the image encoder can be frozen while the user tower is optimized to update user embeddings. Besides resulting in a lighter training, that allows updating user preferences without recomputing image embeddings.

\section{Experimentation}
\label{sec:experiments}

\subsection{Evaluation Metrics}
Evaluation is performed in the withheld test set, which consist of $1M$ clicks among $0.5M$ images and $40k$ users.
Next, we describe the evaluation metrics that are monitored to explore the influence of different features and setups:

\begin{itemize}
\item {\texttt{Accuracy@k}}: Given a test set with users and images they have clicked on, recommendations for those users are predicted. Then it is measured if the clicked image is within the top-k recommendations. Measures the interest of users in the recommendations
\item {\texttt{Catalog Coverage}}: Measures the percentage of test set images appearing in the test set users recommendations. The top-10 recommendations for each user are considered. Monitors the diversity of the recommendations among users.
\item {\texttt{Effective Catalog Size (ECS)}}: Proposed by Netflix \cite{ecs}, measures how diverse is the content shown to users at each k, being k the position in the recommendation ranking. 
It is maximum when all images are recommended evenly.
\item {\texttt{Visual Diversity}}: Average L2 distance within the deep features of top-10 recommendations for each user. Inspired in LPIPS \cite{lpips}, used to measure diversity of GAN generated images. It measures how diverse recommendations are in terms of deep features for each user, which monitors how much the RecSys relies on them to serve recommendations (and implicitly how much it exploits the rest of the features).
\end{itemize}

\subsection{User Features Study}

Table \ref{tab:user} shows the performance metrics of models leveraging all the presented image features but different sets of user features and monitors their influence in recommendations: A model encoding users solely with \texttt{User ID}, a second one using additionaly \texttt{Org ID}, and a third one leveraging also  \texttt{User Country}, \texttt{User Language} and \texttt{Subscription Product}.
It shows that the model benefits from using other features in addition to \texttt{User ID}, the individual user identifier.
Using \texttt{Org ID} together with \texttt{User ID} boosts all metrics, which practically means that users within the same organization share preferred visual styles.
This performance improvement is not due to the fact of just adding more capacity to the model, since adding more capacity to the model leveraging only \texttt{User ID} did not help.

The model using all the presented image features is the one showing superior performance. That proves that, even in a RecSys leveraging unique user identifiers, contextual features help and can be leveraged by simply concatenating feature representations.
The fact that the model leverages \texttt{User Country}, \texttt{User Language} and \texttt{Subscription Product} has another potential benefit: It might be able to provide useful recommendation for users suffering from the cold start problem based on the preferences of users with similar contextual features. 

\begin{table*}[h]
\centering
  \caption{Visual Styles RecSys metrics leveraging different user features and all the presented image features. All features stands for every feature presented in Section \ref{sec:data}.}
  \label{tab:user}
  \begin{tabular}{lccccc}
    \toprule
    Model & Acc@10 & Acc@100 & Catalog Coverage & ECS@10 & Visual Diversity\\
    \midrule
    \texttt{VS RecSys: User ID} & 0.085 & 0.663 & 28.26 & 49719 & 0.303\\
    \texttt{VS RecSys: User ID, Org ID} & 0.112 & 0.801 & 38.62 & 62258 & 0.308\\
    \texttt{VS RecSys: All features} & \textbf{0.144} & \textbf{0.890} & \textbf{51.72} & \textbf{85669} & \textbf{0.309}\\
    \bottomrule
  \end{tabular}
\end{table*}

\subsection{Image Features Study}

Table \ref{tab:deep} shows the influence of different image features in the performance metrics. In this section we analyze how \texttt{Visual Styles RecSys} leverages the presented features to learn user preferences and explain how the proposed set of evaluation metrics enables that model behaviour understanding. The table also includes a popular baseline, which simulates a non-personalized baseline where the top clicked images in the test set are recommended to every user.

\subsubsection{Deep Features Influence}

The model basing recommendations only on \texttt{Deep} features gets a lower score in \texttt{Visual Diversity} than the popular baseline. That's expected: We can consider the \texttt{Visual Diversity} score of the popular baseline as an upper bound, since any ranker leveraging deep features will tend to score lower than a popular ranking which does not. At the same time, we can consider the \texttt{Visual Diversity} score of this model the lower bound, since it bases recommendations entirely on deep features, and any model successfully leveraging other features in addition to \texttt{Deep} features should score higher than this one.

\subsubsection{Shallow Features Influence}

The model leveraging all the presented image features but \texttt{Deep} gets higher \texttt{Visual Diversity} than the one leveraging only \texttt{Deep}, but lower than the popular baseline. That's expected and the reason is that \texttt{Deep} features are still correlated with the shallow features this model bases recommendations on. The \texttt{Accuracy} scores and the general diversity scores (\texttt{Catalog Coverage} and \texttt{ECS}) are lower than for the model based on \texttt{Deep} but still significant. That leads to two conclusions: First that this set of (shallow) features is relevant for recommendations and second that \texttt{Deep} features are key.
However, as discussed before, exploiting \texttt{Deep} features rich representations comes with the risk of biasing recommendations towards the semantics of the projects users have worked on during the training window. To control that, we can monitor the \texttt{Visual Diversity} of the models.

\subsubsection{Joining Deep and Shallow Features}

The model trained with all features scores in \texttt{Visual Diversity} significantly higher than the one leveraging solely \texttt{Deep} features and, as expected, lower than the one not leveraging them. The scores show that it bases recommendations on the combination of \texttt{Deep} features and the rest of the presented ones, as pursued.
It significantly outperforms the rest of the models in \texttt{Accuracy}, \texttt{Catalog Coverage} and \texttt{ECS}. That means that it successfully leverages features of diverse natures to provide recommendations that are of interest to users and diverse among them compared to the other models. In our experimentation, \texttt{Accuracy} and general diversity scores have been always correlated.

Finally, we compare the best \texttt{Visual Styles RecSys} to one where we have replaced \texttt{Deep} features (from the Inception v1 model trained with search results) with \texttt{CLIP} features. 
The drop in \texttt{Visual Diversity} shows that \texttt{CLIP} \cite{clip} features are more discriminative than \texttt{Deep} features and hence the model fits them more, proving also the usefulness of the \texttt{Visual Diversity} metric to monitor the influence of deep features in the recommendations. 
The semantic nature of \texttt{CLIP} features and the low \texttt{Visual Diversity} score evidence that the recommendations of this model are much more based on the semantics of the projects a user has been working on than the ones of the former models. The drop in \texttt{Accuracy} and general diversity metrics shows that our evaluation strategy, which uses the latest $10\%$ of clicks as test set, is also penalizing that.

\begin{figure}[h]
  \centering
  \includegraphics[width=0.5\linewidth]{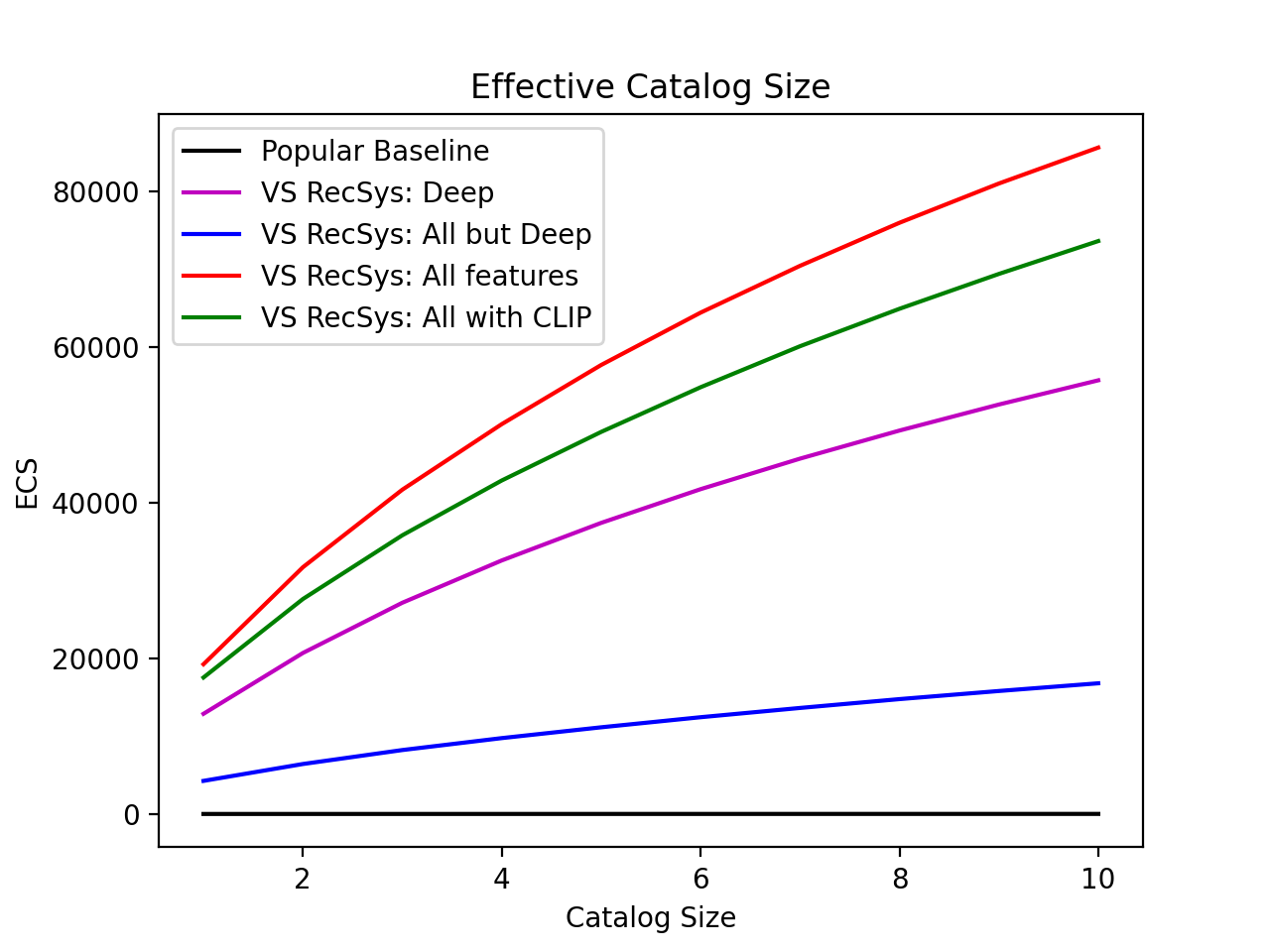}
  \caption{\texttt{Effective Catalog Size} \cite{ecs} of models leveraging different image features and the Popular Baseline. The X axis (Catalog Size) corresponds to the number of recommendations served to each user.}
  \label{fig:ecs}
\end{figure}

Importantly, this means that leveraging all the presented features \texttt{Visual Styles RecSys} has successfully learnt users' preferred visual styles transversal to the projects they work on and, also, that the proposed evaluation strategy is useful to prove that by benchmarking models leveraging different features.
Figure \ref{fig:ecs} shows the \texttt{Effective Catalog Size} at different $k$ of the aforementioned models, confirming that the one leveraging all the presented features provides substantially more diverse recommendations among users.

\begin{table*}[h]
\centering
  \caption{\texttt{Visual Styles RecSys} evaluation metrics leveraging different image features and all the presented user features. Additionaly, a model where \texttt{Deep} features have been substituted by \texttt{CLIP} features, and a baseline where all users get recommended the most popular images.}
  \label{tab:deep}
  \begin{tabular}{lccccc}
    \toprule
    Model & Acc@10 & Acc@100 & Catalog Coverage & ECS@10 & Visual Diversity\\
    \midrule
    \texttt{Popular Baseline} & 0.02 & 0.19 & 2e-6 & 9.4 & \textbf{0.413} \\
    \texttt{VS RecSys: Deep} & 0.093 & 0.655 & 35.75 & 55748 & 0.278\\
    \texttt{VS RecSys: All but Deep} & 0.021 & 0.250 & 14.38 & 16798 & 0.371\\
    \texttt{VS RecSys: All features} & \textbf{0.144} & \textbf{0.890} & \textbf{51.72} & \textbf{85669} & 0.309\\
    \texttt{VS RecSys: All with CLIP} & 0.075 & 0.561 & 32.27 & 73649 & 0.236\\
    \bottomrule
  \end{tabular}
\end{table*}

\subsection{Deep Cross Layers Study}

Cross Layers explicitly model feature interactions, which seems ideal for our image encoder where we aim to learn visual styles.
Table \ref{tab:cross} shows the influence of Deep Cross Layers \cite{dcn2} in the image tower with different configurations: Stacking them after features representations concatenation and before the MLP, and placing them in parallel to the MLP after features representations, to then concatenate the two branches representations and learn the final embedding. The dimensionality of the Cross Layers used is always $1/4$ of the input dimensionality, as recommended in \cite{dcn2}.
The best performing setup is stacking a single cross layer before the MLP, which significantly outperforms a plain MLP.

\begin{table*}[h]
\centering
  \caption{\texttt{Visual Styles RecSys} metrics with different Cross Layers configurations. MLP refers to an image encoder consisting on a Multi Layer Perceptron without Cross Layers. Stacked means that Cross Layers are inserted after image features concatenation. Parallel means that a DCN is placed in parallel to the MLP just after feature concatenation.}
  \label{tab:cross}
  \begin{tabular}{lccccc}
    \toprule
    Model & Acc@10 & Acc@100 & Catalog Coverage & ECS@10 & Visual Diversity\\
    \midrule
    \texttt{VS RecSys: MLP} & 0.121 & 0.840 & 48.15 & 77636 & 0.308\\
    \texttt{VS RecSys: 1 CL stacked} & 0.108 & 0.831 & 45.32 & 71758 & 0.308\\
    \texttt{VS RecSys: 2 CL stacked} & 0.129 & 0.871 & 43.56 & 74391 & 0.301\\
    \texttt{VS RecSys: 1 CL in parallel} & \textbf{0.144} & \textbf{0.890} & \textbf{51.72} & \textbf{85669} & \textbf{0.309}\\
    \texttt{VS RecSys: 2 CL in parallel} & 0.134 & 0.873 & 50.28 & 82311 & 0.306\\    \bottomrule
  \end{tabular}
\end{table*}

\subsection{Qualitative Results}

\begin{figure}[h]
  \centering
  \includegraphics[width=1\linewidth]{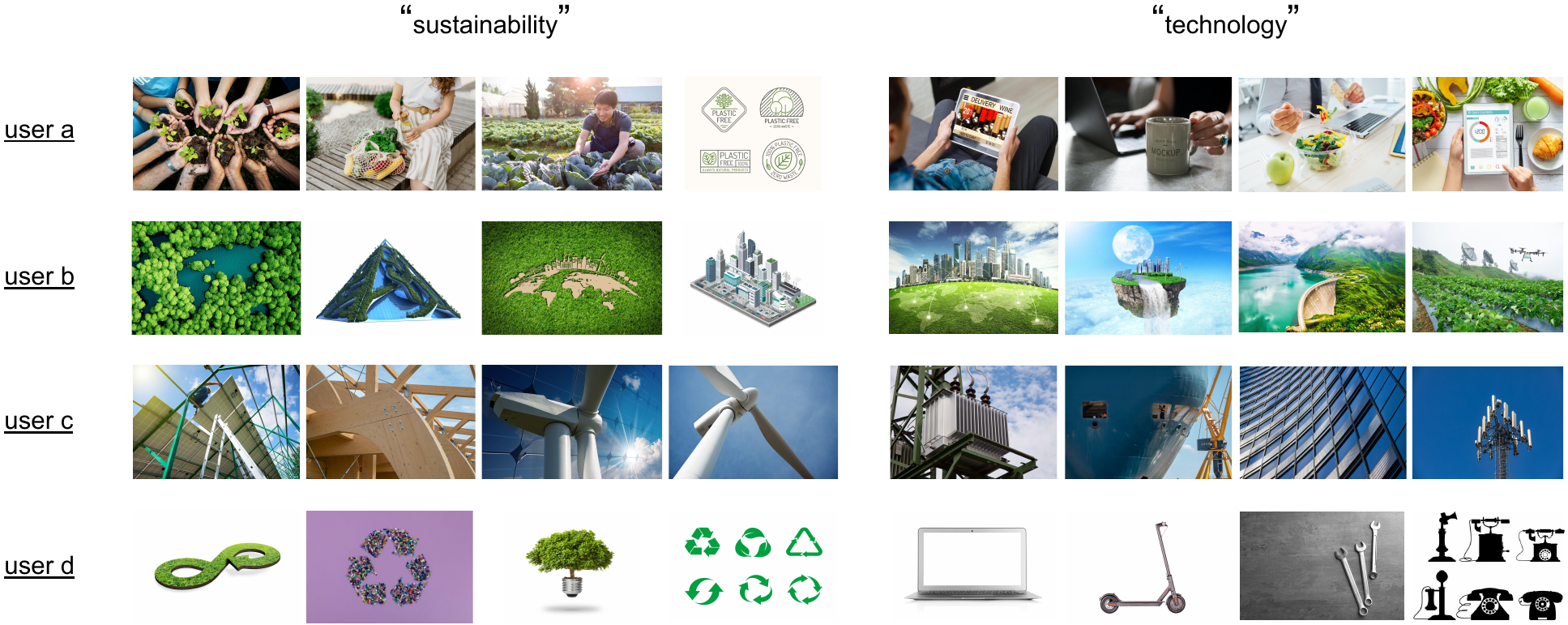}
  \caption{Top ranked images by Visual Styles RecSys for four different users querying for "sustainability" and "technology" images. To better illustrate the learned styles, to get this qualitative results the search engine has been substituted by a simple keyword filtering: For each query, the test set is filtered by the querying keyword, and the images containing that keyword are re-ranked by the RecSys.}
  \label{fig:qualitative_results}
\end{figure}

Figure \ref{fig:qualitative_results} shows qualitative results for different users in the image marketplace demonstrating the diverse styles \texttt{Visual Styles RecSys} can learn. \underline{User a} shows preference towards stocky close-up images featuring people and the gastronomy semantic vertical. Meanwhile, \underline{user b} prefers top-view and wide angle images with greenish colors and digital art. \underline{User c} is more interested in structures close-ups with bluish colors, while \underline{user d} in signs and symbols and isolated objects with homogeneous backgrounds. 

Note that \texttt{Visual Styles RecSys} recommends images based not only on preferred users image features, but on complex styles learned based on the interactions of those features. Additionally, the model does not limit to learn a single preferred visual style for each user, but learns multiple preferred styles and, as a result, recommendations are diverse not only among users but also within them. As examples, in addition to the aforementioned preferences, \underline{user b} has a lighter interest in isolated objects, and \underline{user a} in illustrations.

\subsection{Search Engine and RecSys Integration}
In this section we conduct an experiment to evaluate offline how useful recommendations are to users compared to plain search engine results. We base this evaluation on the top-10 search queries at the image marketplace. For each one of them, we randomly select $200$ users that licensed (bought) one or more images using that search query. Then, using the images licensed as ground truth, we evaluate the Mean Average Precision (\texttt{MAP}) of the first $1000$ plain search engine results and the MAP of re-ranking those results with \texttt{Visual Styles RecSys} for that specific user.
The objective is measuring if re-ranking the top search engine results with \texttt{Visual Styles RecSys} helps showing to users images that are relevant to them first.
Additionally, we evaluate the MAP of two methods that fuse search engine and \texttt{Visual Styles RecSys} rankings: Mean Reciprocal Rank \cite{reciprocal}, which is the average of the respective reciprocal ranks, and Weighted Reciprocal Rank, which is a weighted average of the reciprocal ranks.

\begin{figure}[h]
  \centering
  \includegraphics[width=1\linewidth]{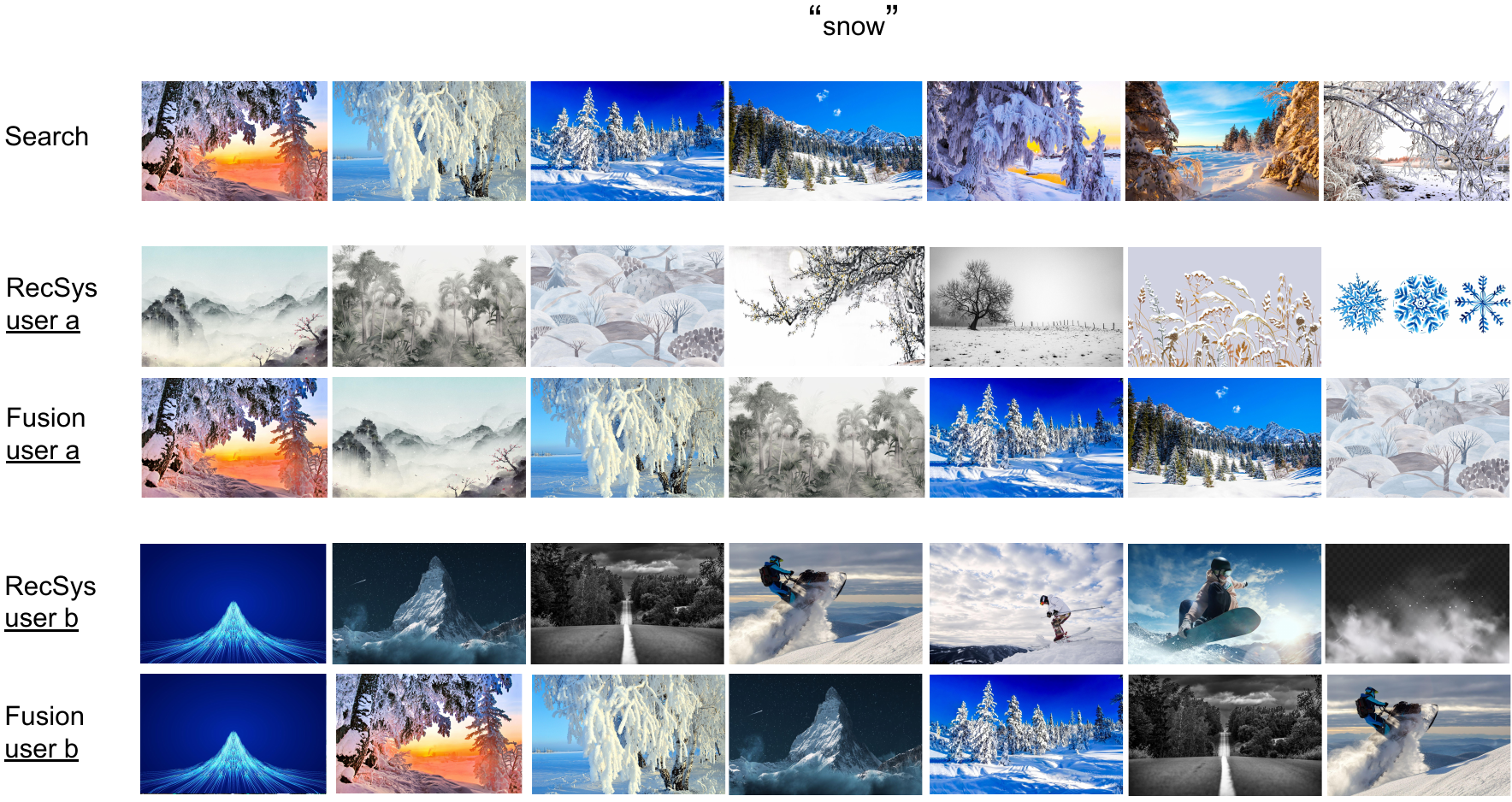}
  \caption{Top search engine, \texttt{Visual Styles RecSys} and Mean Reciprocal Rank results for two different users and the "snow" query. Note that, differently from Figure \ref{fig:qualitative_results}, recommendations have been obtained by re-ranking the top $1000$ search engine results.}
  \label{fig:search_vs_styleranker}
\end{figure}

Table \ref{tab:map} shows the results of the different rankings. \texttt{Visual Styles RecSys} is substantially superior in terms of \texttt{MAP} compared to the plain search engine, which confirms that re-ranking the top search engine results to show first the ones inline with the user' preferred styles is a good strategy. Moreover, fusing both rankings improves performance further, which indicate that images that have a good compromise between popularity and match with user's preferred styles are the most relevant to the user. 
Figure \ref{fig:search_vs_styleranker} shows the top search engine results for the "snow" search query together with the top results of \texttt{Visual Styles RecSys} and Mean Reciprocal Rank fusion for two different users. It illustrates how personalizing search results enhances users' experience. In the example \underline{user a} gets recommended "snow" images with light colors featuring trees and abstract landscapes, close-up views and illustrations, while \underline{user b} gets darker images, mountain landscapes and winter sports images. That might spare them navigating pages of search engine results based mainly on popularity.

\begin{table*}[h]
\centering
  \caption{MAP (Mean Average Precision) scores of the plain search engine results, \texttt{Visual Styles RecSys} and fused rankings based on $100$ user licenses for each one of the top-10 search queries at the image marketplace.}
  \label{tab:map}
  \begin{tabular}{lccccc}
    \toprule
    Ranking & MAP \\
    \midrule
    \texttt{Search Engine} & 0.021 \\
    \texttt{VS RecSys} & 0.030 \\
    \texttt{Mean Reciprocal Rank} & 0.030 \\
    \texttt{Weighted Reciprocal Rank} &  \textbf{0.032}\\
    \bottomrule
  \end{tabular}
\end{table*}

\section{Conclusions}
\label{sec:conclusions}
We have addressed the challenging task of recommending images to creatives in a global content marketplace, where users' interests are linked to the projects they work on and therefore change rapidly and abruptly. 
The proposed solution, \texttt{Visual Styles RecSys}, learns creatives' visual styles preferences transversal to the projects they work on, allowing personalizing search engine results for them while maintaining their ability to use all its functionality for content discovery.
Experimentation has proven how the proposed model successfully leverages features of different natures to learn long term users' style preferences compared to simpler models, and the presented set of evaluation metrics have demonstrated key to understand the behaviour of the benchmarked models by monitoring the influence of different features. 
Specifically, we have seen how a model leveraging all image features but \texttt{Deep} features is not competitive in \texttt{Accuracy} or \texttt{Effective Catalog Size} with models leveraging \texttt{Deep} features, while a model leveraging solely \texttt{Deep} features suffers from \texttt{Visual Diversity}. However, combining shallow and deep features the model achieves superior \texttt{Accuracy} and diversity metrics while maintaining a competitive \texttt{Visual Diversity}.
Nevertheless, if \texttt{Deep} features are replaced with the more discriminative \texttt{CLIP} representations, the model suffers in all metrics given that their rich semantic representations obstruct the model from fitting the rest of the image features and lead to recommendations based on short-term interests. 

The task of learning visual styles preferences transversal to projects in an image marketplace can be equated to other scenarios where one aims to learn recommendations leaving aside short term preferences, and therefore we strongly believe that the presented work will be useful in diverse fields.

\bibliographystyle{unsrt}
\bibliography{main}

\end{document}